# Diffeomorphic Transformer-based Abdomen MRI-CT Deformable Image Registration

Yang Lei[1], Luke A. Matkovic[1], Justin Roper[1], Tonghe Wang[2], Jun Zhou[1], Beth Ghavidel[1], Mark McDonald[1], Pretesh Patel[1] and Xiaofeng Yang[1*]

[1]Department of Radiation Oncology and Winship Cancer Institute, Emory University, Atlanta, GA 30322
[2]Department of Medical Physics, Memorial Sloan Kettering Cancer Center, New York, NY, 10065

**Running title**: MRI-CT Deformable Registration

**Manuscript Type:** Original Research

**Corresponding author:**
Xiaofeng Yang, PhD
Department of Radiation Oncology
Emory University School of Medicine
1365 Clifton Road NE
Atlanta, GA 30322
Tel: (404)-778-8622
Fax: (404)-778-4139
E-mail: xiaofeng.yang@emory.edu

## Abstract

**Background:** Stereotactic body radiotherapy (SBRT) is a well-established treatment modality for liver metastases in patients unsuitable for surgery. Both CT and MRI are useful during treatment planning for accurate target delineation and to reduce potential organs-at-risk (OAR) toxicity from radiation. MRI-CT deformable image registration (DIR) is required to propagate the contours defined on high-contrast MRI to CT images. An accurate DIR method could lead to more precisely defined treatment volumes and superior OAR sparing on the treatment plan. Therefore, it is beneficial to develop an accurate MRI-CT DIR for liver SBRT.

**Purpose:** To create a new deep learning model that can estimate the deformation vector field (DVF) for directly registering abdominal MRI-CT images.

**Methods:** The proposed method assumed a diffeomorphic deformation. By using topology-preserved deformation features extracted from the probabilistic diffeomorphic registration model, abdominal motion can be accurately obtained and utilized for DVF estimation. The model integrated Swin transformers, which have demonstrated superior performance in motion tracking, into the convolutional neural network (CNN) for deformation feature extraction. The model was optimized using a cross-modality image similarity loss and a surface matching loss. To compute the image loss, a modality-independent neighborhood descriptor (MIND) was used between the deformed MRI and CT images. The surface matching loss was determined by measuring the distance between the warped coordinates of the surfaces of contoured structures on the MRI and CT images. To evaluate the performance of the model, a retrospective study was carried out on a group of 50 liver cases that underwent rigid registration of MRI and CT scans. The deformed MRI image was assessed against the CT image using the target registration error (TRE), Dice similarity coefficient (DSC), and mean surface distance (MSD) between the deformed contours of the MRI image and manual contours of the CT image.

**Results:** When compared to only rigid registration, DIR with the proposed method resulted in an increase of the mean DSC values of the liver and portal vein from 0.850±0.102 and 0.628±0.129 to 0.903±0.044 and 0.763±0.073, a decrease of the mean MSD of the liver from 7.216±4.513 mm to 3.232±1.483 mm, and a decrease of the TRE from 26.238±2.769 mm to 8.492±1.058 mm.

**Conclusion:** The proposed DIR method based on a diffeomorphic transformer provides an effective and efficient way to generate an accurate DVF from an MRI-CT image pair of the abdomen. It could be utilized in the current treatment planning workflow for liver SBRT.

# 1. Introduction

Stereotactic body radiation therapy (SBRT) is a high-dose ablative technique administered in one to five fractions, with a biologically effective dose that is either equal to or greater than that of conventional radiotherapy.[1] Recent studies have demonstrated significant local control of primary and metastatic hepatic malignancies through the use of SBRT.[2,3] As a result, SBRT has emerged as a viable alternative for liver-directed therapy in the treatment of both primary liver cancers and liver metastases, particularly for patients ineligible for surgical resection for various reasons.[4,5] However, the liver is a radiation-sensitive organ with a risk for radiation-induced liver dysfunction and hepatic toxicities including exacerbation of any underlying liver cirrhosis with both mortality risk and potential detrimental effect on the quality of life for patients. With high dose per fraction and steep dose gradients seen in SBRT, normal liver dosimetry and optimal delivery of the target dose are hindered by potential hepatic toxicities, which continue to be limiting factors.[6,7]

Simulation CT images are regularly obtained for treatment planning due to their provision of essential electron density information required for dose calculation. By utilizing the spatial precision and administering tumoricidal radiation doses based on the target(s) and OARs defined on planning CT images, SBRT can be implemented to achieve high rates of tumor control while minimizing the exposure of nearby healthy tissue. This approach helps to reduce the risk of radiation-induced liver disease.[8] Although the planning CT image can be used to delineate the target(s) and OARs, the CT image alone may not offer sufficient soft-tissue information to enable accurate contouring of the target(s) and nearby critical structures and relevant organs. This level of accuracy is vital for both tumor control and dose delivery precision.[9]

The integration of MRI into the liver SBRT workflow under free breathing conditions is being studied as a means of enhancing SBRT, thanks to MRI's superior soft tissue contrast.[10] By employing MRI-CT deformable image registration (DIR), contours that were defined on high-contrast MRI images can be transferred to CT images, potentially resulting in smaller treatment volumes, greater OAR sparing accuracy, and reduction in toxicity. Consequently, the development of an accurate MRI-CT DIR for liver SBRT is highly beneficial.

Recent methods have been proposed to train deep learning-based models that can create a transformation from a set of moving and fixed images to a deformation estimation, such as a deformation vector field (DVF).[11] Supervised deep learning-based methods have shown feasibility for DIR.[12] However, such methods require ground truth DVFs for training, which are usually generated through simulation. Such simulation for generating training data may not accurately represent the distribution of real patient deformation. Consequently, recent deep learning-based methods have utilized unsupervised techniques. These methods utilize a network that performs spatial transformations to deform one image into another.[13-

[15] Unsupervised DIR methods consider volumetric changes between scans; however, they may generate unrealistic results where voxels move in a non-physiological manner. This challenge is particularly pronounced in abdominal motion when the DIR is near the diaphragm. In addition to quantitative calculations, qualitative assessment is recommended for DIR.[16] Moreover, the different intensity distributions of distinct image modalities can make liver MRI-CT DIR challenging. Additionally, aligning a specific organ to nearby structures may not provide the best solution for the entire image volume in cases where the organ is deformed relative to adjacent structures. In this study, we aim to develop a novel DIR with a direct approximation of dense DVF via a deep learning-based method, called diffeomorphic transformer-based DIR, to match MRI images to planning CT images and thus improve the accuracy of OAR sparing for liver SBRT. The concepts that inspired us in our methodology are introduced in the following sections.

## 1.A. Probabilistic Diffeomorphic Registration

Several recent studies have assumed that deformations and their inverses are diffeomorphic, or differentiable.[17-22] In one study, deep learning-based probabilistic models were used to apply the diffeomorphic theory for DIR.[18] Diffeomorphic deformations are differentiable and invertible, and thus preserve topology.[19] The approach developed in 2018[20] used a variational strategy to learn a deep learning-based model to predict a stationary velocity field.[21] VoxelMorph,[22] an improved approach developed in 2019, was applied to a multitude of deformable representations and assumed the deformations were diffeomorphic, especially with the stationary velocity field. In this study, we also assumed that the deformations are diffeomorphic. We utilized topology-preserved deformation features extracted from the probabilistic diffeomorphic registration model to accurately obtain abdominal motion, which can be used for estimating the DVF.

## 1.B. Surface-guided

Recent registration techniques have suggested utilizing contour-based loss, such as the Dice similarity coefficient (DSC), in place of image-based similarity terms when contours are accessible through optimization for intra-subject multi-modality.[23] Instead of relying on volume-based similarity metrics, surface matching methods employ surface coordinates or geometric features extracted from anatomical structures to evaluate similarity.[24] One solution is to use iterated closest point-based optimization methods to find the shape correspondences.[25] The method in this work combined surface- and image-based losses to train a deep learning-based model, which utilized a 3D point representation with volumetric images to achieve fast registration. The registration was enabled by a differentiable surface distance function.

## 1.C Transformer

Recently, vision transformer architectures have been proposed to overcome the limitations of convolutional neural networks (CNNs) and have produced state-of-the-art performances in many medical imaging applications.[26] Transformers can be strong candidates for image registration because their substantially larger receptive field enables more precise comprehension of the spatial correspondence between moving and fixed images. In this work, we aimed to integrate the transformer architecture into a CNN for deformation estimation. The Swin Transformer,[27] which demonstrated superior performance in motion tracking, was integrated into the model for deformation feature extraction.

## 2. Methods and materials

### 2.A. Mathematics

Let $I_{MR}$ and $I_{CT}$ be the 3D MRI and CT images, where MR is taken as the moving image and CT is taken as the fixed image, and let $\varphi \in R^3 \to R^3$ denote the DVF that deformably registers the MRI to match the CT. The deformed image $I_{def}$ is described as $I_{MR}$ warped via $\varphi * I_{MR}$ to match $I_{CT}$. Inspired by previous works,[20,22] the estimated $\varphi$ yields diffeomorphic registration given $I_{MR}$ and $I_{CT}$ in a probabilistic manner. Let $y_\varphi$ denote the displacement field, which is estimated from the proposed CNN-based model and registers $I_{MR}$ by $\varphi * I_{MR}$. The prior probability of $y_\varphi$ can be assumed to have a multivariate Gaussian distribution, $P(y_\varphi) = \mathcal{N}(y_\varphi; \mu_{y_\varphi}, \sigma_{y_\varphi})$, where the mean is $\mu_{y_\varphi}$ and the covariance is assumed to be $\sigma_{y_\varphi}$. Inspired by a recent study,[17] $y_\varphi$ is assumed to be a fixed velocity field that defines a diffeomorphism through the ordinary differential equation.[28] Namely, the spatial smoothness of $y_\varphi$ is encouraged by $\sigma_{y_\varphi}^{-1} = \lambda L_G$, where $L_G = D_G - A$ is a Laplacian of a neighborhood graph defined on the voxel grid, $D_G$ is the graph degree matrix, and $A$ is a voxel neighborhood adjacency matrix. $\lambda$ represents a parameter that governs the magnitude of the velocity field $y_\varphi$. The posterior distribution of $I_{def}$ can be represented as:

$$P(I_{def}|y_\varphi; I_{MR}) = \mathcal{N}(I_{def}; \varphi * I_{MR}, \sigma_{def}) \quad (1)$$

where $\sigma_{def}$ captures the variance of additive image noise.

Based on the above assumption, the goal of the proposed CNN-based model was to estimate the most likely displacement field $y_\varphi$ for an image set $(I_{MR}, I_{CT})$ which can satisfy a maximum a posteriori estimation $P(I_{def}|y_\varphi; I_{MR})$. Namely, the goal of the CNN-model was the estimation of the voxel-wise velocity field mean $\mu_\varphi$ and variance $\sigma_{y_\varphi}$. The computation of $P(I_{def}|y_\varphi; I_{MR})$ is inflexible, but can be approximated by minimizing the Kullback–Leibler (KL) divergence:

$$\min_{\varphi} KL\left(P_\varphi(I_{def}|y_\varphi; I_{MR}) \parallel P(I_{def}|y_\varphi; I_{MR})\right). \quad (2)$$

As a result, the computation of the negative value of the variational lower bound for the model evidence is achieved. Then, $y_\varphi$ can be learned by optimizing Eq. (2) via stochastic gradient methods. Namely, for a training image pair $(I_{MR}, I_{CT})$, we compute $\varphi * I_{MR}$ with the resulting loss:

$$L(y_\varphi; I_{MR}, I_{CT}) = KL\left(P_\varphi(I_{def}|y_\varphi; I_{MR}) \parallel P(I_{def}|y_\varphi; I_{MR})\right) \tag{3}$$

The minimization of Eq. (3) requires an image similarity loss calculated between $I_{def}$ and $I_{CT}$. Since $I_{def}$ is the deformed MRI image and a different modality than $I_{CT}$, we used our recently well-developed modality-independent neighborhood descriptor (MIND) to compute this cross-modality image similarity loss.[29] More details of the benefit and technical description of MIND loss are discussed in our previous work.[30] Generally, MIND relies on the similarity of small image patches within a single image and seeks to extract distinctive structures in local neighborhoods that are preserved across modalities. It has the capability to differentiate between various features, including corners, edges, and homogeneously textured regions. The multi-dimensional descriptor can be calculated efficiently in a dense manner across the entire image, offering point-wise local similarity across modalities by considering the absolute or squared difference between descriptors. The sum of squared differences in MIND representations of images served as the similarity metric employed for MIND loss.

While in the training phase, the proposed model used delineations of structures of interest on both CT and MRI scans. These contours were incorporated in the loss functions during the training process. Notably, once the training was complete, the model no longer required the CT- and MRI-delineated contours as input; it solely relied on the CT and MRI scans as input. As illustrated in Fig. 1, only the portions indicated by the black arrows were essential for the feedforward path of the model. Specifically, for deformable registration of a new MRI and CT pair using the trained model, only these specified portions were necessary.

## 2.B. Surface-based semi-supervision

Certain training images may provide additional anatomical outlines for specific structures of interest, which can improve the registration in an unsupervised manner. Given the anatomical structure (liver) delineated from $I_{MR}$ and $I_{CT}$, the anatomical surface was extracted. Let $s_{MR}$ and $s_{CT}$ indicate the spatial coordinates of the anatomical structure's surface in $I_{MR}$ and $I_{CT}$, respectively. Given the diffeomorphism $y_\varphi$ in the previous section, we modeled surface location $s_{MR}$, which was formed by displacing a matching surface location $s_{CT}$ according to $y_\varphi$:

$$P(s_{def}|y_\varphi; s_{MR}) = \mathcal{N}(s_{CT}; \varphi * s_{MR}, \sigma_s) \tag{4}$$

where the composition $\varphi * s_{MR}$ warps surface coordinates, $s_{def}$ denotes the warped coordinates of $s_{MR}$, and $\sigma_s$ denotes the spatial variance. The structure (liver) delineation in CT is required only during the training phase. After training, the model no longer requires the delineated contour for a new set of CT and MRI images.

By leveraging both images and contour maps during the training process, our objective was to feed surfaces of the contours of $I_{CT}$, called $s_{CT}$, into the model to approximate the conditional posterior probability $P(y_\varphi | I_{CT}, s_{CT}; I_{MR}, s_{MR})$. As in the previous section, our goal was to minimize the KL divergence between the true posterior and the approximate posterior:

$$\min_\varphi KL\left(P_\varphi(I_{def}|y_\varphi; I_{MR}) \parallel P(y_\varphi | I_{CT}, s_{CT}; I_{MR}, s_{MR})\right) \quad (5)$$

The KL divergence, also known as relative entropy, is a measure of how one probability distribution diverges from a second, expected probability distribution. In this work, it is mathematically defined as:

$$\begin{aligned}&KL\left(P_\varphi(I_{def}|y_\varphi; I_{MR}) \parallel P(y_\varphi | I_{CT}, s_{CT}; I_{MR}, s_{MR})\right) \\ &= \sum_i P_\varphi(I_{def}|y_\varphi; I_{MR}; i) \times \log \frac{P_\varphi(I_{def}|y_\varphi; I_{MR}; i)}{P(y_\varphi | I_{CT}, s_{CT}; I_{MR}, s_{MR}; i)}\end{aligned} \quad (6)$$

where $P_\varphi$ denotes the probability distribution of the ground truth segmentation mask and $P$ denotes the probability distribution of the predicted segmentation mask generated by the model. The sum is taken over all possible outcomes (pixels) $i$ in the distribution.

The goal of KL divergence is to make the predicted mask $P$ as close as possible to the ground truth mask $P_\varphi$. To achieve this, for each pixel $i$, the KL divergence computes the difference in probability assigned by the model $P(y_\varphi | I_{CT}, s_{CT}; I_{MR}, s_{MR}; i)$ and the actual probability in the ground truth $P_\varphi(I_{def}|y_\varphi; I_{MR}; i)$. The summation over all pixels provides a measure of how well the predicted mask aligns with the ground truth at each spatial location. During training, the objective is to minimize the KL divergence between the predicted and ground truth distributions. This encourages the model to learn accurate and spatially aligned segmentations.

## 2.C. Workflow

The framework, called the diffeomorphic transformer model, is summarized in Fig. 1. The first part of the network, called the morphological transformer, takes the images and surfaces as input, then generates the estimated posterior probability, which can be represented by the estimated DVF's mean $\mu_{y_\varphi}$ and variance $\sigma_{y_\varphi}$. To generate a diffeomorphic deformation field $y_\varphi$, a velocity field $y$ is sampled and transformed using integration layers that support differentiability. At last, a spatial transformer warps $I_{MR}$ and $s_{MR}$ to derive deformed images, represented by $\varphi * I_{MR}$ and $\varphi * s_{MR}$, that can match $I_{CT}$ and $s_{CT}$.

To enable optimization of parameters $y_\varphi$ using Eq. (3) and Eq. (5), $I_{def} = \varphi * I_{MR}$ needs to be formed. Given $y_\varphi = \mu_{y_\varphi} + \sqrt{\sigma_{y_\varphi}} \cdot n_v$, where $n_v \sim \mathcal{N}(0,1)$,[31] the $\varphi$ can be computed as $\varphi = e^{y_\varphi}$. The vector integration layer is composed of scaling and squaring operations. Specifically, scaling and squaring operations involve compositions within the neural network architecture using a differentiable spatial transformation operation. Given two 3D vector fields $a$ and $b$, for each voxel $p$ this operation computes $(a \circ b)(p) = a(b(p))$, a non-integer voxel location $b(p)$ in $a$, using linear interpolation. Starting with $\varphi^{1/2^T} = p + y_\varphi/2^T$, we computed $\varphi^{1/2^{t-1}} = \varphi^{1/2^t} \circ \varphi^{1/2^t}$ recursively $T$ times resulting in $\varphi = e^{y_\varphi}$.[32] In the final step, we employed a spatial transform layer to deform the moving image $I_{MR}$ based on the estimated deformation vector field.

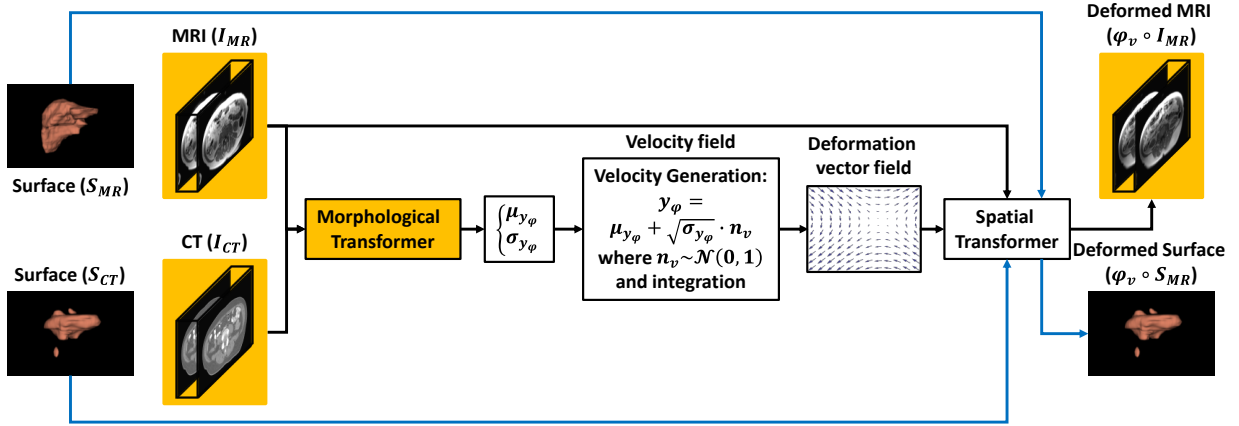

**Figure 1.** The workflow of the proposed diffeomorphic transformer model. The blue arrows denote supervision during training stage, which is not required during inference stage. Both black and blue arrows are needed during training.

In summary, the network takes images $I_{MR}$ and $I_{CT}$ as input, estimates parameters $\mu_{y_\varphi}$ and $\sigma_{y_\varphi}$ used for obtaining the DVF, samples a new velocity field $y_\varphi \sim \mathcal{N}\left(\mu_{y_\varphi|I_{MR},I_{CT}}, \sigma_{y_\varphi|I_{MR},I_{CT}}\right)$, generates a diffeomorphic $\varphi$, and deforms $I_{MR}$. As all the steps' optimization objectives are differentiable, the network parameters can be optimized via stochastic gradient descent optimization. This network-based model yields three outputs, namely $\mu_{z|I_{MR},I_{CT}}$, $\sigma_{z|I_{MR},I_{CT}}$, and $\varphi * I_{MR}$, which are used in the model loss in Eq. (5).

With the optimized network, we can perform the deformable registration of a set of scans $(I_{MR}, I_{CT})$ using $\varphi$. The first step is to obtain the most probable velocity field $\hat{y}_\varphi$ by employing the following equation:

$$\hat{y}_\varphi = \arg\max_{y_\varphi} P\big(y_\varphi|I_{CT}, S_{CT}; I_{MR}, S_{MR}\big) = \mu_{y_\varphi|I_{MR},I_{CT}} + \sqrt{\sigma_{y_\varphi|I_{MR},I_{CT}}} \cdot n_v \qquad (7)$$

by evaluating the proposed CNN. Then, we computed $\varphi$ by utilizing the integration process based on scaling and squaring.

## 2.D. Morphological Transformer

In this work, we developed a morphological transformer, which is a hybrid transformer-CNN model that is able to utilize abdominal MRI-CT ($I_{MR}, I_{CT}$), for $\mu_{y_\varphi|I_{MR},I_{CT}}$ and $\sigma_{y_\varphi|I_{MR},I_{CT}}$ estimation. The morphological transformer is an end-to-end encoder-decoder network, where Swin transformers were employed as the

encoder to capture the spatial correspondence between the input MRI and CT images. The Swin transformer used in this work is a shift-window based transformer, which is inserted to each of the corresponding blocks in Fig. 2. The source code is based on "Swin3D" https://github.com/microsoft/Swin3D.

Then, a CNN-based decoder processed the information provided by the encoder into a deformation feature map. The Conv3D blocks included two 3D convolutional layers. The Conv3D layer was followed by batch normalization and then ReLU activation. Long skip connections were implemented to preserve the flow of localization information between the encoder and decoder stages. Finally, the morphological transformation estimated the dense DVF and applied the DVF to deform the MRI image to match the CT image.

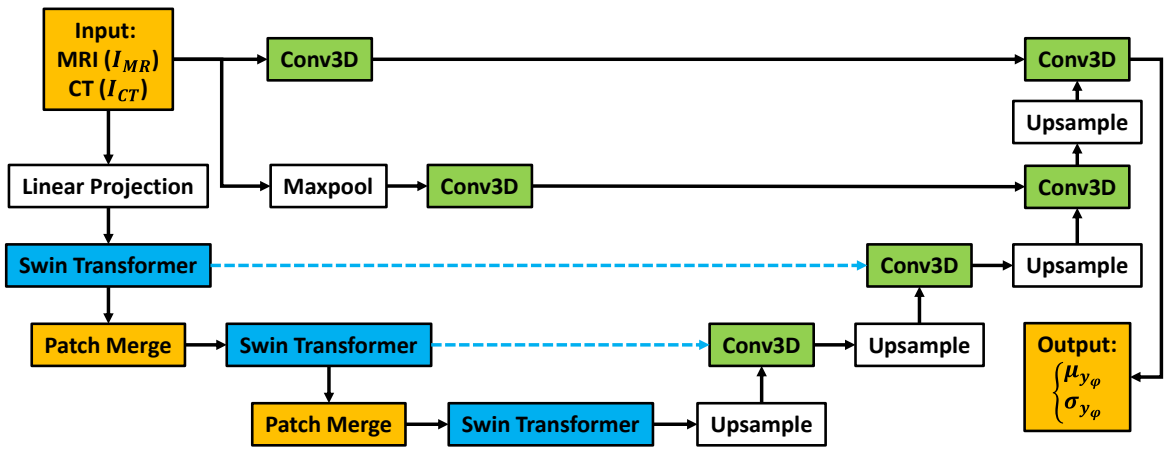

**Figure 2.** Network architecture of the morphological transformer. The blue dashed arrow indicates the skip connection between operators, while the black solid arrows signify the flow to the next set of operators.

## 2.E. Dataset, implementation and evaluation

To test the proposed method, a retrospective study was conducted on a cohort of 50 patients' datasets. These individuals were patients who had liver cancer and underwent radiotherapy. The MRI images were acquired via Axial T1 fast spoiled gradient-echo (FSPGR) scanning. The spatial resolutions of MRI and CT images were resampled to 0.97×0.97×3.00 mm$^3$ prior to DIR. The liver and portal vein structures were manually contoured on MRI and CT images separately.

The investigated deep learning networks were designed using Python 3.6 and TensorFlow and implemented on a GeForce RTX 2080 GPU that had 12 GB of memory. To reduce the computational cost, the input data was cropped within the body. The input size was 256×256×128. The convolution kernels employed in each layer of Conv3D measured 3×3×3, with the number of kernels set at 6, 12, 24, and 36 across various hierarchy stages. As for the Swin Transformer, the stages were configured at 3, the embedding size was established at 128 after three stages, sinusoidal position embedding was used, the window size was set to 7×7×5, and the depth configuration was [2, 4, 9] across 3 stages. After linear projection, the patch size of the Swin transformer was 16×16×8. Optimization was performed using the Adam gradient optimizer. The drop path rate was set to 0.2. The learning rate was 2×10$^{-4}$. With the batch size setting of 8 during training

(entire 3D as input), the percentage of utility of GPU was 97%. The implemented network comprised a total of 46.7 million parameters. The training of the model took 15 hours. Once the network was trained, it only took approximately 3 minutes for DVF estimation.

The performance of the proposed method was evaluated using a five-fold cross-validation approach. In detail, the 50 datasets were initially divided randomly and evenly into five groups. Four of these groups were utilized for training purposes, while the remaining group was reserved for testing. This process was repeated five times, with each group serving as the testing set in a rotation.

Qualitative evaluations of the proposed method were performed by visually assessing the alignment between the planning CT and deformed MRI images. A fusion image between the planning CT and deformed MRI images was generated for visual assessment. Quantitatively, the deformed MRI image was evaluated against the CT image using the TRE, DSC, and MSD calculated between the deformed contours of the MRI image and the manual contours of the CT image.

## 3. Results

### 3.A. Comparison with state-of-the-art

Two state-of-the-art deep learning-based methods[13,30] were compared in this work. VoxelMorph formulated DIR as a function that maps an input image pair to a deformation field that aligns these images. A CNN was used to parameterize the mapping function. Given a new pair of scans, VoxelMorph generated a DVF by directly evaluating the function. In MIND, a deep learning-based network was proposed to directly predict the DVF for MRI-CT liver DIR. To overcome the challenge of multimodal registration, a modality-independent descriptor was incorporated into the deep learning network to explore the correlations between the MRI and CT images. The difference of the proposed method when compared to the two recent deep learning-based methods can be summarized as: (1) rather than using unsupervised learning as in the other methods, the proposed method was a semi-supervised method in which liver contour surface matching was used as a contour-based loss to supervise the model, and (2) the proposed method incorporated the Swin transformer into a CNN-based model, the utility of which was demonstrated in the previous subsection.

The deformation results are shown in Fig. 3. Row (a) shows three axial slices of the planning CT image of a different patient, where the corresponding original MRI images are shown in row (b). Row (c) shows the fusion images of the planning CT (a) and the original MRI images (b). Row (d) shows the deformed MRI image via VoxelMorph.[13] Row (e) shows the fusion images between the planning CT (a) and deformed MRI images (d). Row (f) shows the deformed MRI images via MIND.[30] Row (g) shows the fusion images between the planning CT (a) and deformed MRI images (f). Row (h) shows the deformed MRI images using the proposed method with the Swin transformer. Row (i) shows the fusion images between the planning CT (a) and deformed MRI images (h).

Several observations can be drawn from Fig. 3. MIND had mismatches for some OARs with higher contrast, whereas the other two methods performed well. For example, when comparing (c1) to (d1) and (e1), MIND was found to produce an enlarged liver and reduced spleen. VoxelMorph didn't work well for cases with abdominal compression when compared to the other methods. When comparing (c2) to (d2) and (e2), one can see that the deformed images of the proposed method and the MIND method show agreement in the CT (fixed image) at the compressed abdominal region, whereas VoxelMorph does not. Finally, the proposed method shows better bowel matching, which can be seen when comparing (c3) to (d3) and (e3).

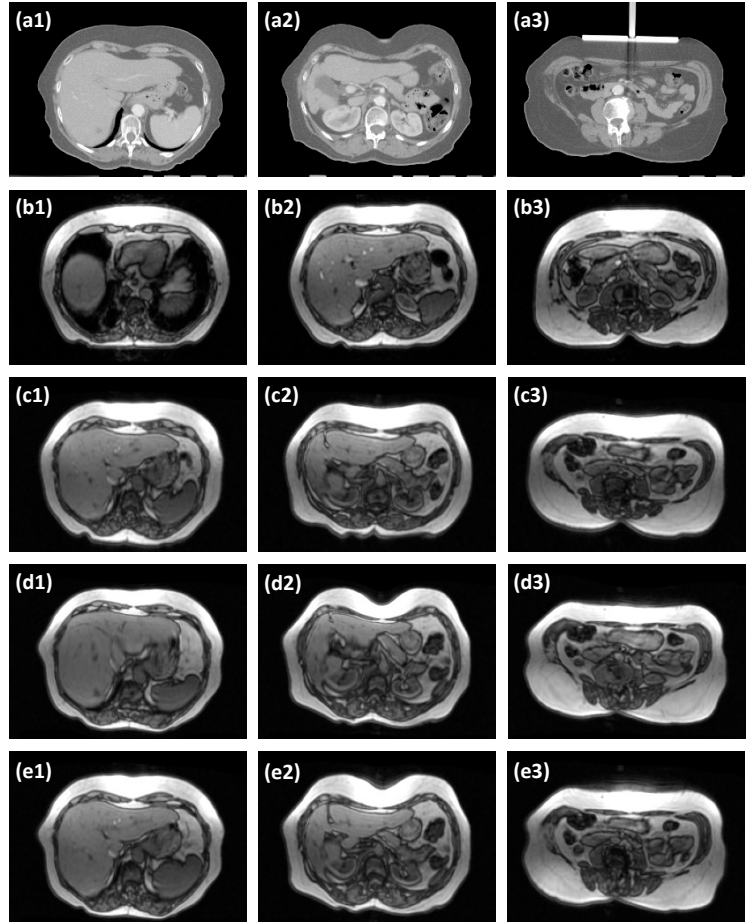

**Figure 3.** Row (a): planning CT images, which are regarded as fixed images. Row (b): MRI images, which are regarded as moving images. Row (c): deformed MRI images using VoxelMorph.[13] Row (d): deformed MRI images using MIND.[30] Row (e): deformed MRI images using the proposed method incorporating the Swin transformer.

Table 1 shows the numerical results. Overall, the mean DSC values of the liver and portal vein increased from 0.850±0.102 and 0.628±0.129 to 0.903±0.044 and 0.763±0.073 after DIR using the proposed method. The mean MSD of the liver decreased from 7.216±4.513 mm to 3.232±1.483 mm. The TRE decreased from 26.238±2.769 mm to 8.492±1.058 mm . In terms of TRE, the proposed method outperforms the other two methods by about 10%.

Table 1. Overall quantitative results achieved via the proposed method and state-of-the-art methods. P-values with each method evaluated against the proposed method are shown below the mean±std.

|  | DSC of liver | DSC of portal vein | MSD of liver (mm) | TRE (mm) |
|---|---|---|---|---|
| Before DIR | 0.850±0.102 (p=0.001) | 0.628±0.129 (p<0.001) | 7.216±4.513 (p<0.001) | 26.238±2.769 (p<0.001) |
| VoxelMorph | 0.894±0.057 (p=0.346) | 0.742±0.116 (p=0.194) | 3.730±1.911 (p=0.122) | 9.364±1.043 (p<0.001) |
| MIND | 0.889±0.068 (p=0.198) | 0.752±0.131 (p=0.550) | 3.519±1.833 (p=0.368) | 10.775±1.356 (p<0.001) |
| Proposed | 0.903±0.044 | 0.763±0.073 | 3.232±1.483 | 8.492±1.058 |

## 3.B. Ablation study

To demonstrate the utility of the Swin transformer used in the proposed diffeomorphic deformation model, the performances with and without integration of the Swin transformer into the proposed network were compared using identical environmental settings. To compare these two methods fairly, in the network without the Swin transformer, the Swin transformer was replaced by CNN layers with equalized numbers of learnable parameters. The deformation results are shown in Fig. 4. Row (a) shows three axial slices of the planning CT image of the same patient. Row (b) shows the corresponding slices of the MRI image. Row (c) shows the deformed MRI using the proposed method without incorporating the Swin transformer. Row (d) shows the deformed MRI using the proposed method with incorporation of the Swin transformer.

Three observations can be drawn from Fig. 4: (1) the abdominal motion, specifically in the bowel region, liver-lung interface region, and stomach region, would be enlarged between the planning CT and MRI images as shown in row (b); (2) both methods perform well on higher contrast OARs, such as the kidneys and spinal cord; (3) at the dorsal region, the proposed method can show more reasonable deformation when incorporating the Swin transformer than without, which can be seen from the comparison between row (c) (deformed MRI image without transformer) and row (d) (deformed MRI image with transformer), especially when observing (c2) and (c3) vs (d2) and (d3).

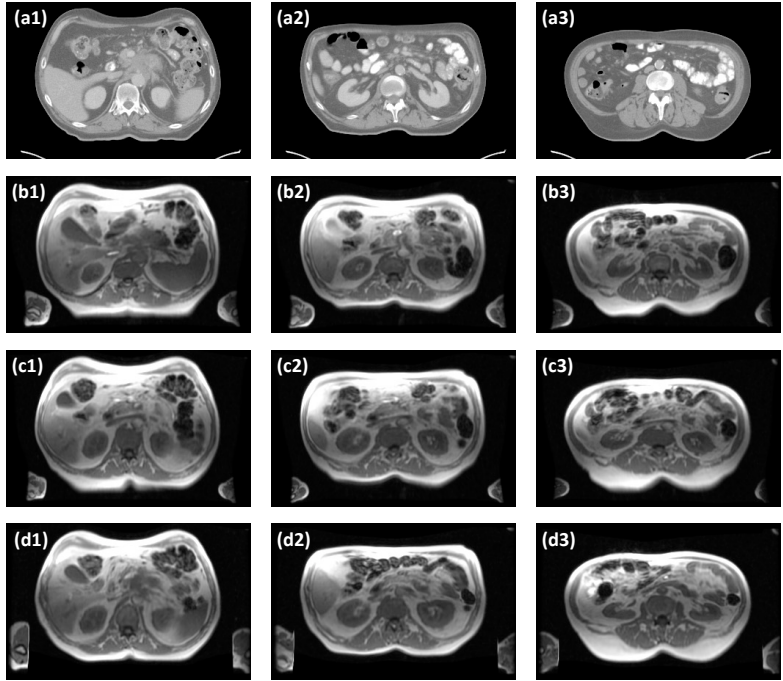

**Figure 4.** Row (a): planning CT images, which are regarded as fixed images. Row (b): MRI images, which are regarded as moving images. Row (c): deformed MRI images using the proposed method without incorporating the Swin transformer. Row (d): deformed MRI images using the proposed method incorporating the Swin transformer.

The average TREs over all patients are listed in Table 2. The average TRE is 8.492±1.058 mm for the proposed method incorporating the Swin transformer, which is better than the proposed method without incorporating the Swin transformer. The smaller standard deviations for the TREs of the proposed method also demonstrate better registration robustness. Table 2 shows that by incorporating the Swin transformer, the DSCs for both the liver and portal vein of the deformed contour improved when compared to the planning CT image's contour.

**Table 2.** Overall quantitative results achieved with and without the Swin transformer in the proposed method. Reported p-values compare the use of the transformer vs without the transformer and before DIR.

|  | DSC of liver | DSC of portal vein | MSD of liver (mm) | TRE (mm) |
|---|---|---|---|---|
| Before DIR | 0.850±0.102 | 0.628±0.129 | 7.216±4.513 | 26.238±2.769 |
|  | (p=0.001) | (p<0.001) | (p<0.001) | (p<0.001) |
| Without Transformer | 0.898±0.047 | 0.740±0.118 | 3.422±1.846 | 9.026±1.044 |
|  | (p=0.614) | (p=0.220) | (p=0.615) | (p=0.018) |
| With Transformer | 0.903±0.044 | 0.763±0.073 | 3.232±1.483 | 8.492±1.058 |

The utilized MIND loss represents a local mutual information characterization of cross-modality images. For cross-modality image registration, the normalized cross-correlation (NCC) loss is another viable option. NCC serves as a similarity metric by measuring the correlation between two images while accounting for variations in intensity. NCC loss encourages the deep learning model to find a transformation that maximizes the normalized cross-correlation between the reference and target images. To evaluate the performance of MIND as compared to NCC loss, we incorporated a comparison with the proposed network trained using the NCC loss instead of the MIND loss.

The numerical evaluations across all patients are detailed in Table 3. The average TRE for the proposed method employing the NCC loss is 10.154±0.993 mm, indicating a less favorable outcome compared to the proposed method using the MIND loss. Additionally, contour alignment metrics such as DSC and MSD for the model trained with NCC loss are also lower than those achieved with MIND loss, as illustrated in Table 3. This discrepancy is primarily attributed to the challenge faced by the NCC loss in addressing the significant intensity differences between MRI and CT data.

Table 3. Overall quantitative results achieved using MIND loss and NCC loss in the proposed method.

|  | DSC of liver | DSC of portal vein | MSD of liver (mm) | TRE (mm) |
|---|---|---|---|---|
| MIND | 0.903±0.044 | 0.763±0.073 | 3.232±1.483 | 8.492±1.058 |
| NCC | 0.870±0.054 | 0.668±0.157 | 3.500±1.491 | 10.154±0.993 |
|  | ($p<0.001$) | ($p<0.001$) | ($p=0.260$) | ($p<0.001$) |

## 4. Discussion

In this study, we presented a new approach for liver SBRT MRI-CT cross-modality DIR using a diffeomorphic transformer-based method. Our method assumed diffeomorphic deformations and leveraged topology-preserved deformation features extracted from a probabilistic diffeomorphic registration model to accurately capture abdominal motion and estimate the DVF. To enhance the deformable feature extraction, we integrated Swin transformers, which have shown excellent performance in motion tracking, into a CNN-based model. This integration allowed us to extract high-quality features that capture the deformations accurately. We optimized our model using a combination of volume-based similarity for unsupervised training and surface matching for semi-supervised training. This dual optimization approach ensures that the generated DVF not only aligns the volumes but also matches the surfaces with a particular focus on OARs. By increasing the reasonability of the generated DVF, we aimed to improve the overall quality of the registration process. Paired two-sample t-tests (two-tailed) were used to evaluate the significance. For Table 1, the proposed method has better metrics than VoxelMorph and MIND for the DSCs of the liver and portal vein, MSD of the liver, and TRE, with associated p-values shown. Compared to VoxelMorph and MIND, TRE is the metric that is statistically significant ($p<0.001$) for the proposed method. The differences for DSCs and MSD were not statistically significant, however the mean and standard deviation values for the proposed method are improved over both VoxelMorph and MIND in all cases. For Table 2, all values of mean and standard deviation are improved with the transformer than without, excluding the standard deviation for the TRE, with statistically significant results for TRE.

As compared to deterministic registration methods, the probabilistic DIR accounts for spatial variability in deformations. Instead of providing a single deterministic transformation, it can offer a distribution of likely transformations, providing a more comprehensive representation of the possible spatial variations. To achieve this in our work, as shown in Eq. (5) and Eq. (6), the proposed method optimized the model based on the likelihood function, which is the KL divergence between the true segmentation and the approximate segmentation. The benefit of this framework is the capability of dealing with inherent ambiguities between the MRI and CT images, especially when dealing with complex anatomical structures, such as in the abdominal region.

Three potential issues exist with the current proposed method. First, the performance of DIR can be affected by the MRI image quality. As can be seen from Fig. 3, the patient's MRI image quality is affected by inhomogeneity. Our optimization of the model during training was based on MIND, which is a cross-modality image similarity metric, and surface matching loss. Although surface matching loss may not be affected by this involved bias, MIND, which is based on the texture information compared between deformed MRI and planning CT images, can be affected by this bias. To reduce this potential issue, incorporating MRI bias correction as a first step into our deep learning-based model will be a future focus. Recently, generative adversarial networks (GANs) have been used for MRI intensity non-uniformity correction.[33] However, the computation complexity of the GAN can be dramatically increased due to involvement of an additional discriminator. A more efficient method of integrating MRI image quality improvement into our deep learning-based model can be another future focus.

Another potential limitation is that the model trained by MIND image similarity loss may not be able to cover the full range of MRI modalities. In this work, we only trained and tested our model on T1-weighted MRI with FSPGR scanning. Clinically, the patient may not have been scanned using T1-weighted MRI. In a future work, inclusion of variable MRI modalities to train our model will be another focus.

The landmarks used for evaluating the proposed method and comparing it to other methods are the same for all methods and were selected from a region of large abdominal motion and the surface of the liver. By using the proposed method, the TRE decreased from 26.238±2.769 mm before DIR to 8.492±1.058 mm after DIR. When compared to recent studies of multi-modality liver deformable registration,[34] our TRE performance is larger (as compared to mean landmark errors of 3.20-5.36 mm in [34]). One major reason may come from the different datasets. In our work, the slice thickness was 3 mm, whereas in [34] the mean slice thicknesses were 1.64 mm for T2 MRI and 0.8 mm for T1 MRI. Another potential reason may be that the multi-modality MRI sequences used in [34], such as contrast-enhanced T1, T2, and diffusion-weighted imaging MRI, provide more comprehensive structural information when compared to only using T1 in our work. Testing the model's performance on multi-modality MRI sequences for liver DIR will be another future focus.

## 5. Conclusion

We developed a deformable image registration method for abdomen MRI-CT images using a diffeomorphic transformer. The method estimates a DVF from an MRI-CT image pair and applies it to deform the MRI image to match the CT image. This provides an effective solution for registering abdominal MRI-CT images, which can be useful in delineating target volumes and OARs for liver SBRT.

**Acknowledgments**

This research is supported in part by the National Cancer Institute of the National Institutes of Health under Award Number R56EB033332, R01EB032680 and P30CA008748.